\documentclass[letterpaper]{article} 
\usepackage{aaai25}  
\usepackage{times}  
\usepackage{helvet}  
\usepackage{courier}  
\usepackage[hyphens]{url}  
\usepackage{graphicx} 
\usepackage{natbib}  
\usepackage{caption} 
\usepackage{algorithm}
\usepackage{algorithmic}

\usepackage{newfloat}
\usepackage{listings}
\DeclareCaptionStyle{ruled}{labelfont=normalfont,labelsep=colon,strut=off} 
\floatstyle{ruled}
\newfloat{listing}{tb}{lst}{}
\floatname{listing}{Listing}
\usepackage{amsmath}
\usepackage{amssymb}
\usepackage{multirow}
\usepackage{booktabs}
\usepackage{colortbl}
\usepackage{subcaption}

\title{Promptable Representation Distribution Learning and Data Augmentation for Gigapixel Histopathology WSI Analysis}
\author{
    Kunming Tang\textsuperscript{\rm 1,\rm 2},
    Zhiguo Jiang\textsuperscript{\rm 1,\rm 2,\rm 3},
    Jun Shi\textsuperscript{\rm 4},
    Wei Wang\textsuperscript{\rm 5,\rm 6},
    Haibo Wu\textsuperscript{\rm 5,\rm 6},
    Yushan Zheng\textsuperscript{\rm 1}\thanks{Corresponding author}
}
\affiliations{

    \textsuperscript{\rm 1}Beijing Advanced Innovation Center on Biomedical Engineering, School of Engineering Medicine, Beihang University\\
    \textsuperscript{\rm 2}Image Processing Center, School of Astronautics, Beihang University\\
    \textsuperscript{\rm 3}Tianmushan Laboratory\\
    \textsuperscript{\rm 4}School of Software, Hefei University of Technology\\
    \textsuperscript{\rm 5}Department of Pathology, the First Affiliated Hospital of USTC\\
    \textsuperscript{\rm 6}Intelligent Pathology Institute, Division of Life Sciences and Medicine, University of Science and Technology of China\\
    \{tkm2000,jiangzg,yszheng\}@buaa.edu.cn, juns@hfut.edu.cn, weiwang@hmfl.ac.cn, wuhaibo@ustc.edu.cn
}

\usepackage{bibentry}

\begin{document}

\maketitle

\begin{abstract}
Gigapixel image analysis, particularly for whole slide images (WSIs), often relies on multiple instance learning (MIL).
Under the paradigm of MIL, patch image representations are extracted and then fixed during the training of the MIL classifiers for efficiency consideration.
However, the invariance of representations makes it difficult to perform data augmentation for WSI-level model training, which significantly limits the performance of the downstream WSI analysis.
The current data augmentation methods for gigapixel images either introduce additional computational costs or result in a loss of semantic information, which is hard to meet the requirements for efficiency and stability needed for WSI model training.
In this paper, we propose a Promptable Representation Distribution Learning framework (PRDL) for both patch-level representation learning and WSI-level data augmentation. Meanwhile, we explore the use of prompts to guide data augmentation in feature space, which achieves promptable data augmentation for training robust WSI-level models.
The experimental results have demonstrated that the proposed method stably outperforms state-of-the-art methods. 
\end{abstract}

%
\begin{links}
    \link{Code}{https://github.com/lazytkm/PRDL}
\end{links}

\section{Introduction}
\label{sec:intro}
Histopathology whole slide image (WSI) classification~\cite{lu2021data, shao2021transmil, yang2022remix, zheng2023kernel} is increasingly popular in computer-aided pathological diagnosis, presenting unique challenges to the field of computer vision~\cite{chen2022scaling, nakhli2023sparse, zhang2022gigapixel}. 
Unlike conventional natural images, WSIs have massive image resolutions, often reaching up to billions of pixels. 
To tackle the gigapixel problem, a variety of methods within this domain employ multiple instance learning (MIL) frameworks ~\cite{ilse2018attention, li2021dual, campanella2019clinical, zhang2022dtfd} to address the specific needs of WSI analysis.

\begin{figure*}[tb]
  \centering
  \includegraphics[width=0.9\linewidth]{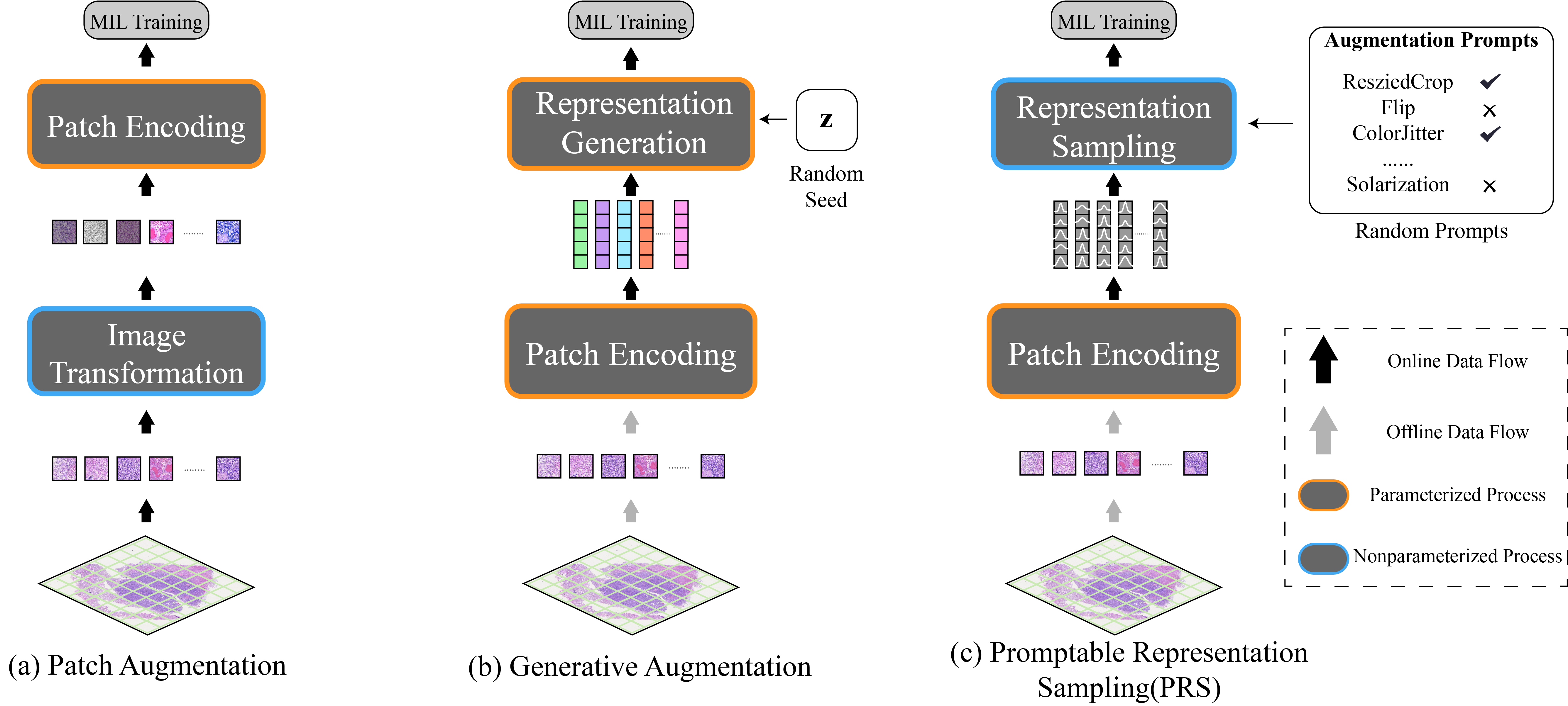}
  \caption{Comparison between existing methods for WSI data augmentation and our method. (a) represents the traditional image augmentation used in natural images, which is inefficient. (b) involves the use of generative models for data augmentation in feature space. (c) describes our promptable representation sampling strategy tailed for WSI augmentation.}
  \label{fig:compare}
\end{figure*}

Within the MIL framework, WSI analysis is usually divided into three phases: 1) Divide a WSI into patches; 2) Extract features for these patches; and 3) Aggregate these features to make a prediction for the entire WSI.
Data augmentation serves as an effective expanding data strategy for training deep models ~\cite{chen2020simple}. 
It is also important in the domain of histopathology WSI analysis. 
As shown in Figure \ref{fig:compare}(a), if we draw a parallel to the process used for natural images, data augmentation for WSI should ideally be performed continuously after the WSI patching stage and before the patch-level feature extraction stage. 
However, for efficiency, the first two stages are generally conducted only once during the entire training process ~\cite{lu2021data, shao2021transmil}.
This leads to the inapplicability of traditional image augmentation techniques, thereby inspiring the shift towards performing data augmentations directly in feature space.

\begin{figure*}[tb]
  \centering
  \includegraphics[width=0.85\textwidth]{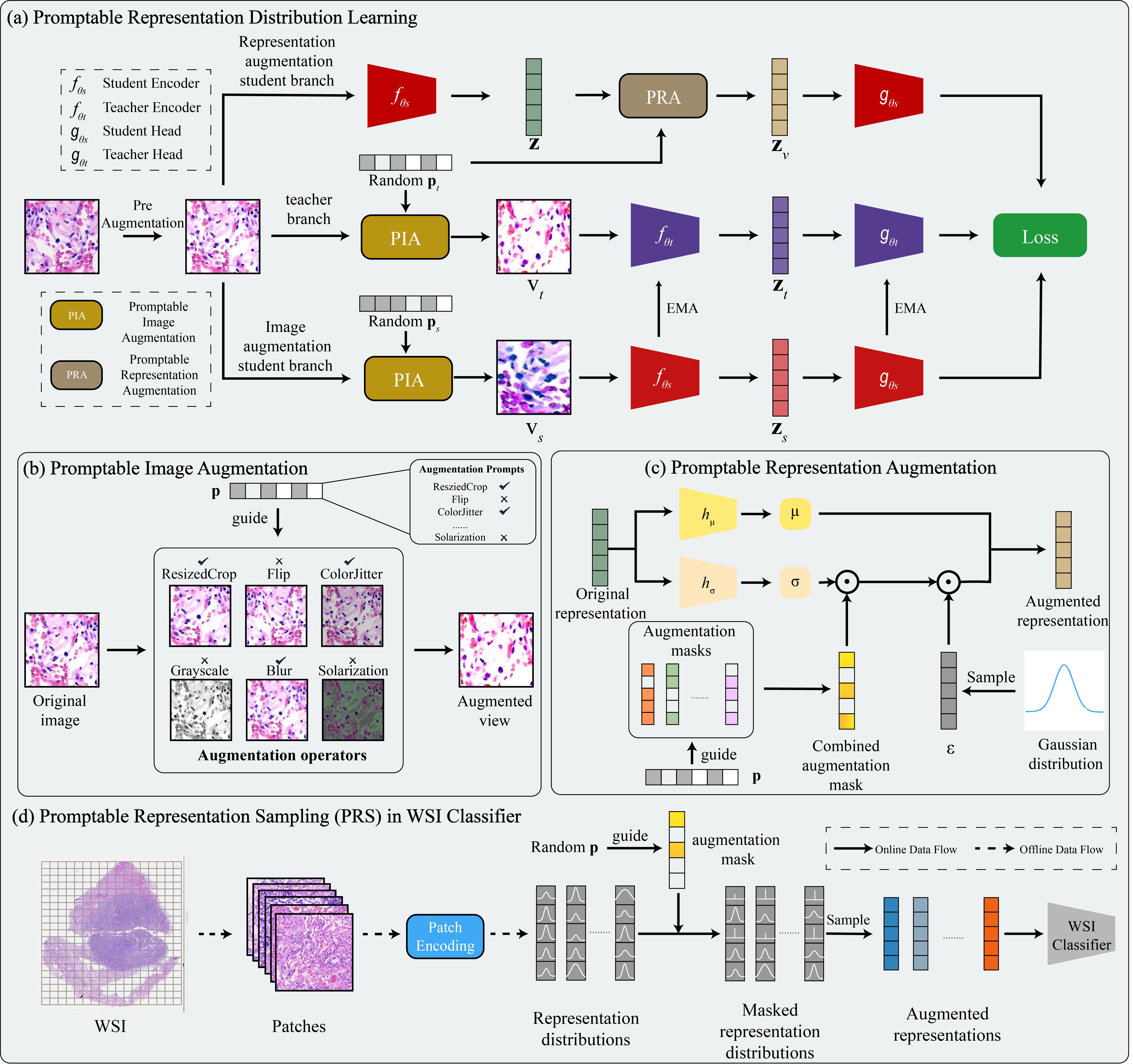}
  \caption{The proposed representation learning and WSI data augmentation framework includes (a) the process of PRDL, where the two student branches share weights in the encoder and head, (b) and (c) provide detailed descriptions of the modules in (a), and (d) shows the flowchart of WSI augmentation during training.
  }
  \label{fig:method}
\end{figure*}

To achieve data augmentation for WSI, several studies have proposed using generative methods ~\cite{dai2024augdiff,zaffar2022embedding} or Mixup techniques  ~\cite{yang2022remix, chen2023rankmix,gadermayr2023mixup} to create data augmentations in feature space, as illustrated in Figure \ref{fig:compare}(b).
However, these methods either introduce additional computational costs due to the need for training another parameterized model or cause a loss of semantic information.
Additionally, the augmented results generated by these methods often lack control, in contrast to data augmentations applied directly in image space where changes are more visually intuitive and easier to manage.
Currently, there lacks data augmentation methods to meet the unique demands of gigapixel image analysis that are not only computationally efficient but also preserve control. 

In this paper, we introduce a novel approach named promptable representation distribution learning (PRDL) to address the challenges associated with patch-level representation learning and WSI-level data augmentation in histopathological WSI analysis.
Within this framework, a representation distribution estimator is designed and trained during self-supervised representation learning.
As shown in Figure \ref{fig:compare}(c), this estimator is capable of predicting a distribution of potential representation augmentations for each patch. 
After patch-level feature extraction, each patch within a WSI is represented as an individual distribution rather than a static point in feature space.
Additionally, the representation distribution of each dimension is confined within a range by specific augmentation prompts, to simulate different augmentation operations that are used in image space.
Finally, we implement a non-parameterized representation augmentation process through online sampling patch representations from these prompted distributions, to efficiently achieve the data augmentation for WSI-level model training.
The proposed method was evaluated on a lung dataset with 754 WSIs and two public lung datasets with 696 WSIs and 3064 WSIs.
The experimental results have demonstrated that the proposed method stably outperforms state-of-the-art methods.
The main contributions of this paper can be summarized as follows:

(1) We proposed promptable representation distribution learning (PRDL), a novel representation learning framework with prompted representation distribution estimation for WSI classification. 
A promptable distribution estimator is designed to incorporate representation augmentation into representation learning.
Compared with the traditional image-level data augmentation, the proposed method can provide more expansive augmentation. 
This significantly improves the discrimination of the patch representations and thereby enhances the performance of the subsequent WSI analysis model.

(2) We designed a promptable representation sampling (PRS) module based on PRDL. 
Utilizing PRS, we successfully facilitated the interchange between data augmentation and patch encoding processes, and achieved promptable and flexible data augmentation in the feature space for gigapixel histopathology image analysis.
Furthermore, we leveraged the augmentation prompts in image space to guide the training of the learnable augmentation masks in feature space.
This strategy enables us to conduct representation augmentation with greater control, enhancing the flexibility of the augmentation process.

\section{Related Work}
\subsection{Data Augmentation for WSI analysis}
The conventional method of data augmentation for WSI, similar to that used for natural images, involves continuously extracting representations of augmented patches from each "bag" (a set of patches) throughout the training process.
However, this is obviously inefficient for WSI model training due to the huge amount of time required for feature extraction.
Consequently, augmentation is primarily performed during WSI preprocessing, as seen in methods like ABMIL~\cite{ilse2018attention}, to enhance patch diversity before training.

Data augmentation strategies developed for WSI analysis fall into two main categories.
The former is realized with generative models~\cite{zaffar2022embedding,dai2024augdiff}, while the latter generates new subsets from bags through various mixing approaches~\cite{yang2022remix,chen2023rankmix, gadermayr2023mixup}.
Specifically, DAGAN ~\cite{zaffar2022embedding} involves training networks like generative adversarial networks to create synthetic data augmentations within the feature space.
However, these generative models require an extra training phase separate from the representation model, as well as additional computational resources during the inference stage.
ReMix~\cite{yang2022remix} mixes class-specific prototypes determined by K-means clustering.
Although Mixup methods can enhance data diversity, they may sometimes generate representations that deviate from the distribution of real-world data, which potentially compromises the model's performance on actual datasets.
The shared shortcoming of these data augmentation methods is their inability to direct the augmentation process, leading to a lack of control. 

\subsection{Self-Supervised Representation Learning}
In MIL methods, it is essential to select an appropriate patch encoder. 
CNN models pre-trained on the ImageNet ~\cite{russakovsky2015imagenet} often serve as patch encoders due to their robust feature extraction capabilities ~\cite{lu2021data, shao2021transmil}.
However, there are inevitably semantic differences between pathological and natural images.
Self-supervised learning (SSL) has been widely used in representation learning ~\cite{jaiswal2020survey,bao2021beit,huang2023mast,assran2023self,zhou2021ibot}. 
Chen et al. ~\cite{chen2020simple} propose an end-to-end model SimCLR and systematically study the impact of data augmentations.
They observe that single augmentation is not sufficient to learn good representations.
MoCov2 ~\cite{chen2020improved} expands upon MoCo ~\cite{he2020momentum} by incorporating blur augmentation, thereby improving the baseline on ImageNet.
Grill et al.~\cite{grill2020bootstrap} develop BYOL, a unique metric-learning approach that learns representations by predicting one view from another, demonstrating the importance of color diversity in augmentations.
Most recently, DINO ~\cite{caron2021emerging} is proposed to utilize self-supervised ViT~\cite{dosovitskiy2020image} for representation learning. 
DINO ~\cite{caron2021emerging} follows the data augmentations of BYOL ~\cite{grill2020bootstrap} and multi-crop ~\cite{caron2020unsupervised}, which have also proven the advantage for patch representation pretraining in histopathology WSI analysis ~\cite{chen2022scaling, wu2024pan} .
Our approach leverages the DINO framework and ViT architecture, setting this combination as the baseline for our representation learning. 
Unlike previous methods that focus on image space, we investigate the potential of applying data augmentation in feature space to enhance representation learning.

\section{Method}
Our goal is to develop a data augmentation strategy that operates within the feature space after the encoding of image patches, rather than applying traditional image augmentation techniques before patch encoding.
Additionally, it is crucial that the representation augmentation remains as promptable as traditional image augmentations, avoiding that the outcomes are unpredictable and harm the performance.

As illustrated in Figure \ref{fig:method}a, the architecture of our proposed model is constructed on the foundation of DINO~\cite{caron2021emerging}.
We extend the model by introducing an additional student branch specifically for representation augmentation. This branch shares weights with the image augmentation student branch and incorporates a promptable representation augmentation module.
Within this framework, we can employ prompts aligned with image augmentation to guide the augmentation in feature space.

Following the training of the model, as depicted in Figure \ref{fig:method}d, we can continuously obtain diverse WSI data in feature space through online sampling representations from the distributions.
These distributions can be adjusted by different combinations of the trained augmentation masks.
In this way, we can perform data augmentations in feature space that are as promptable as image augmentations.

\subsection{Promptable Representation Distribution Learning}
In self-supervised learning, the typical approach involves generating two different sets of augmented data from the same original data to help the model learn useful features without labeled data. 
In this approach, we modify traditional image augmentation by integrating a set of prompts that specify the type of augmentation.
Given an image, we obtain $\Tilde{\textbf{x}} \in \mathbb{R}^{w \times h \times 3}$ by pre-augmentation, including flip, color distortion and gray scaling.
Assuming that we have an augmentation set $\mathcal{T}$ consisting of $K$ augmentation operators $o_1,\dots,o_K$, we produce two augmentation prompts 
$\textbf{p}_t \in \{0,1\}^K$ and $\textbf{p}_s \in \{0, 1\}^K$ by sampling random compositions of augmentation operators, where a value of $1$ appears in the $i$-th bin of the prompt indicates the $i$-th augmentation operator $o_k$ is active.

\subsubsection{Promptable Image Augmentation}
Image augmentation is the basis of traditional self-distillation. 
We first define the augmentations required in our image augmentation student branch and teacher branch.
Guided by the augmentation prompts $\textbf{p}_t$ and $\textbf{p}_s$, we produce two different views $\textbf{v}_t = t(\Tilde{\textbf{x}}|\textbf{p}_t)$ and $\textbf{v}_s = t(\Tilde{\textbf{x}}|\textbf{p}_s)$.
The views $\textbf{v}_t$ and $\textbf{v}_s$ are further encoded by $f_{\theta_t}$ and $f_{\theta_s}$ to obtain their representations $\textbf{z}_t = f_{\theta_t}(\textbf{v}_t)$ and $\textbf{z}_s = f_{\theta_s}(\textbf{v}_s)$.
Subsequently, these representations are transformed into embeddings through projector heads $g_{\theta_t}$ and $g_{\theta_s}$.
Referring to the knowledge distillation paradigm, we train the student network to match the output of the given teacher network, parameterized by $\theta_t$ and $\theta_s$, respectively.

\subsubsection{Representation Distribution Estimation}
The core of the representation augmentation student branch is representation distribution estimation. 
In our method, we construct a neural network that can be trained to act as an estimator, using Gaussian prior to characterizing the distribution of patch representations ~\cite{zang2021fasa}.
This architecture includes two main components: a mean head $h_{\mu}$ and a variance head $h_{\sigma}$, both composed of fully connected layers.
Given the inherent non-negativity of the variance, we compute its logarithm rather than the variance itself directly.
The corresponding representation mean ${\boldsymbol{\mu}} \in \mathbb{R}^D$ and standard deviation ${\boldsymbol{\sigma}} \in \mathbb{R}^D$ of each image that together define a Gaussian distribution $\mathcal{N}(\boldsymbol{\mu}, \boldsymbol{\sigma}^2)$, which can be calculated by equations:
\begin{equation}
    \boldsymbol{\mu} = h_{\mu}(f_{\theta_s}(\Tilde{\textbf{x}})),\quad \boldsymbol{\sigma} = \exp{(h_{\sigma}(f_{\theta_s}(\Tilde{\textbf{x}})) / 2)}.
\end{equation}

\subsubsection{Promptable Representation Augmentation}
The estimated distribution $\mathcal{N}(\boldsymbol{\mu}, \boldsymbol{\sigma}^2)$ handles the potential outcomes of representation augmentations for the patch. To make the distribution reflect representations from specific augmentation operators, we introduce a set of masks $K$ denoted by $\textbf{M}=[\textbf{m}_1,\textbf{m}_2,...,\textbf{m}_K]^{\mathrm{T}}$, where each row $\textbf{m}_k$ corresponds to a specific augmentation operator $o_k$ from the set of transformations $\mathcal{T}$.
To ensure $\textbf{M} \in (0, 1)^{K \times D}$, we adopt a strategy where $\textbf{M}$ is obtained by applying the sigmoid function to another random initialized matrix  $\textbf{U} \in \mathbb{R}^{K \times D}$.
By incorporating the augmentation prompt $\textbf{p}_t$, we constrain this distribution to a more specific augmentation space. 
Preciously,
\begin{equation}
    \boldsymbol{\sigma}_{p_t} = \boldsymbol{\sigma} \odot \textbf{m}_{p_t}, \quad \textbf{m}_{p_t} = \textbf{p}_t\textbf{M} / \|\textbf{p}_{t}\|_1,
\end{equation}
where $\odot$ represents Hadamard product.
With the narrowed distribution $\mathcal{N}(\boldsymbol{\mu}, \boldsymbol{\sigma}_{p_t}^2)$ for a patch, we can obtain variable representations of the patch by sampling process $\textbf{z}_v \sim \mathcal{N}(\boldsymbol{\mu}, \boldsymbol{\sigma}_{p_t}^2)$, which is computationally efficient. 
Furthermore, by configuring $\textbf{p}_t$, we can control the specific type of augmentation to be executed.
Here, we adopt the reparameterization trick~\cite{doersch2016tutorial} to enable backpropagation for training.
Specifically, we sample representations $\textbf{z}_v$ under the representation independence assumption
\begin{equation}
    \textbf{z}_v = \boldsymbol{\mu} + \boldsymbol{\sigma}_{p_t} \odot \boldsymbol{\epsilon},\quad \boldsymbol{\epsilon} \sim \mathcal{N}(0, \textbf{I}_D),
\end{equation}
where $\textbf{I}_D$ denotes $D$-dimensional identity matrix.

\begin{figure*}[tb]
  \centering
  \includegraphics[width=\linewidth]{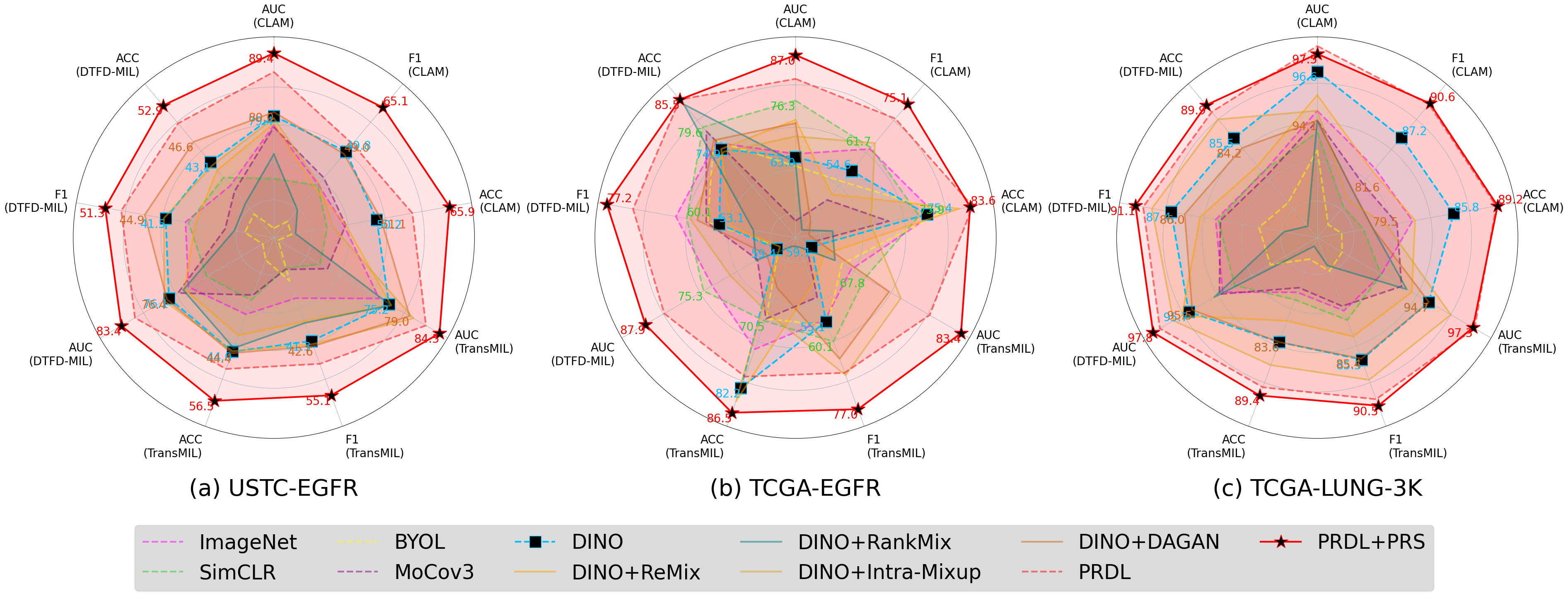}
  \caption{Comparisons with SOTA Methods. Please refer to the supplemental material in the extended version for complete numerical results.
  }
  \label{fig:chart}
\end{figure*}

\subsection{Objective and Optimization}
\subsubsection{Knowledge Distillation }
Then, we adapt the augmentation paradigm to self-supervised learning.
Following the structure of DINO ~\cite{caron2021emerging}, we obtain the probability distribution over $D$ dimensions denoted by $\textbf{y}_t$ by a softmax operation on the branch outputs:
\begin{equation}
    {\textbf{y}_t}=softmax(g_{\theta_t}(\textbf{z}_t)/\tau_t) \label{q},
\end{equation}
where $\tau_t > 0$ is a temperature parameter that controls the sharpness of the probability distribution. Similar formulas hold for $\textbf{y}_s$ and $\textbf{y}_v$ from the two student branches with the mapping head $g_{\theta_s}$.
We minimize the basic loss function 
\begin{equation}
\mathcal{L}_{CE}=H(\textbf{y}_t, \textbf{y}_s) +  H(\textbf{y}_t, \textbf{y}_v),
\end{equation}
where $H(\textbf{a},\textbf{b})=-\textbf{a} \log \textbf{b}$.
It is important to note that we assign the same random prompt $\textbf{p}_t$ for a patch to obtain $\textbf{z}_t$ and $\textbf{z}_v$. 
It guides the distillation architecture to align representations from both the image augmentation and representation augmentation for the same combination of augmentation operators. 
This is the basis on which we can decouple the augmentation operators and thereby control the process of the representation augmentation.
Moreover, we follow the DINO~\cite{caron2021emerging} framework to adopt the multi-crop strategy by using 2 global views and several local views.
All crops are fed into the image augmentation student branch while only the global views are fed into the teacher branch.

\subsubsection{Representation Distribution Constraint}
As we model the representation distribution based on Gaussian prior, We add a Kullback-Leibler (KL) divergence constraint to the distribution estimator, which is represented as
\begin{equation}
    \mathcal{L}_{KL}=D_{KL}(\mathcal{N}(0, \textbf{I}_D) | \mathcal{N}(\boldsymbol{\mu}, \boldsymbol{\sigma}^2)),
\end{equation}
where $\mathcal{N}$ is the Gaussian distribution and $D_{KL}$ is the K-L divergence.

\subsubsection{Promptable Mask Constraint}
The prompted masks aim to identify the most specific feature dimensions associated with different augmentations. Therefore, we employ L1 normalization to induce sparsity in the augmentation masks:
\begin{equation}
    \mathcal{L}_{sp}=\Vert \textbf{m}_{p_t} \Vert_1,
\end{equation}
Additionally, we introduce a variance regularization term on the standard deviation of the embeddings across the feature dimension ~\cite{bardes2021vicreg}, to mitigate the issue of augmentation masks trending towards zero:
\begin{equation}
    \mathcal{L}_{var}=max(0, 1 - \sqrt{Var(\textbf{m}_{p_t}) + \gamma}),
\end{equation}
where $\gamma$ is a small scalar preventing numerical instabilities.

\subsubsection{Overall Objective}
The final object function is composed as follows
\begin{equation}
    \mathcal{L}_{total}=\mathcal{L}_{CE} + \beta_1\mathcal{L}_{KL} + \beta_2\mathcal{L}_{sp} + \beta_3\mathcal{L}_{var},
\end{equation}
where $\beta_1, \beta_2$ and $\beta_3$ controls the weights of each term in the loss. 
The parameters of the student network and the distribution estimator are optimized by the gradient descent algorithm, and the teacher network is updated by the exponential moving average (EMA) mechanism
\begin{equation}
    \theta_t \gets \lambda \theta_t + (1 - \lambda) \theta_s
\end{equation}
with the update rule of $\lambda$ following a cosine schedule during training.

\begin{table*}
    \footnotesize
    \centering
    \begin{tabular}{l|ccc|ccc|ccc}
        \toprule
        \multirow{2}{*}{\textbf{Methods}} & \multicolumn{3}{c|}{\textbf{CLAM}~\cite{lu2021data}} & \multicolumn{3}{c|}{\textbf{TransMIL}~\cite{shao2021transmil}}& \multicolumn{3}{c}{\textbf{DTFD-MIL}~\cite{zhang2022dtfd}}\\
        & AUC & F1-Score & ACC &
        AUC & F1-Score & ACC &
        AUC & F1-Score & ACC \\
        \midrule
        DINO+Random Perturbation &
        74.9 & 35.0 & 40.8 &
        74.1 & 39.2 & 39.9 &
        75.6 & 42.8 & 43.9 \\
        DINO+MC Sampling &
        75.9 & 37.2 & 39.5 &
        75.2 & 36.7 & 38.6 &
        81.3 & 45.0 & 47.1 \\
        \midrule
        PRDL w/o PRA (DINO) & 
        79.4 & 49.8 & 50.2 &
        75.2 & 41.3 & 44.0 &
        76.1 & 41.5 & 43.1 \\
        PRDL w/o $\mathcal{L}_{KL}$ &
        83.8 & 51.2 & 51.6 &
        81.0 & 43.1 & 46.6 &
        80.0 & 39.3 & 41.7 \\
        PRDL w/o $\mathcal{L}_{sp}$ &
        83.8 & 52.7 & 54.3 &
        78.4 & 43.5 & 45.3 &
        79.6 & 41.7 & 44.0 \\
        PRDL w/o $\mathcal{L}_{var}$ &
        84.6 & 49.9 & 51.6 &
        80.8 & 43.1 & 49.8 &
        80.9 & 39.6 & 47.3 \\
        \midrule
        \rowcolor[gray]{.8}
        PRDL &
        86.4 & 52.5 & 57.9 &
        82.0 & 47.1 & 48.4 &
        81.3 & 48.7 & 49.8 \\
        \rowcolor[gray]{.8}
        PRDL+PRS &
        \textbf{89.4} & \textbf{65.1} & \textbf{65.9} &
        \textbf{84.5} & \textbf{55.1} & \textbf{56.5} &
        \textbf{83.4} & \textbf{51.3} & \textbf{52.9} \\
        \bottomrule
    \end{tabular}
    \caption{Ablation study of the proposed framework on the USTC-EGFR Dataset.}
    \label{table:ablation}
\end{table*}

\subsection{WSI Analysis with Representation Augmentation}
Given a WSI $\textbf{X} \in \mathbb{R}^{W \times H \times 3}$ sized by $W \times H$, consisting of patches $(\textbf{x}_1,\textbf{x}_2,\dots.\textbf{x}_{n})$,
a common MIL model for WSI classification can be formulated as:
\begin{equation}
{\hat{Y}}=\psi_{\theta_2}(f_{\theta_1}(\textbf{x}_1),f_{\theta_1}(\textbf{x}_2),\dots,f_{\theta_1}(\textbf{x}_{n}))
\end{equation}
where $\hat{Y}$ is the WSI-level prediction, $f_{\theta_1}$ is the patch encoder and $\psi_{\theta_2}$ is the WSI-level representation aggregator. 

\subsubsection{Promptable Representation Sampling}
Through knowledge distillation in the promptable representation distribution learning, we achieved the goal of procedure exchange of data augmentation and patch encoding.
Then we proposed a representation augmentation strategy named promptable representation sampling (PRS) for WSI augmentation.
As illustrated in Figure \ref{fig:method}d, patch-level representation distributions are obtained after patch encoding.
During WSI classifier training, patch representations $\textbf{Z} = (\textbf{z}_1, \textbf{z}_2, \dots, \textbf{z}_n)$ within $\textbf{X}$ can be replaced by representations online sampled from the corresponding distributions with random combined augmentation prompts, which can be formulated as
\begin{equation}
\textbf{Z}_v=({\textbf{z}}_{v_1}, {\textbf{z}}_{v_2}, \dots, {\textbf{z}}_{v_n}), \quad {\textbf{z}}_{v_i} \sim \mathcal{N}(\boldsymbol{\mu}_i, \boldsymbol{\sigma}^2_i).
\end{equation}
Then, we feed $\textbf{Z}_v$ instead of $\textbf{Z}$ into the WSI model to achieve data augmentation for training, which is represented as
\begin{equation}
{\hat{Y}}=\psi_{\theta_2}({\textbf{z}}_{v_1}, {\textbf{z}}_{v_2}, \dots, {\textbf{z}}_{v_n}),
\end{equation}

As in natural images, WSI augmentation is only used in the training phase to improve the generalization of the model.
The image augmentation student branch and the representation augmentation student branch share weights in patch-level representation learning, determining that the WSI augmentation module can be skipped in the inference phase.
Therefore, we can directly feed representations extracted from the original patches into the WSI classifier for prediction:
\begin{equation}
{\hat{Y}}=\psi_{\theta_2}({\textbf{z}_1},{\textbf{z}_2},\dots,{\textbf{z}_n}).
\end{equation}

\section{Experiments}
The proposed method was evaluated based on WSI classification tasks in three datasets:
1) \textbf{USTC-EGFR}: a private dataset with 754 WSIs categorized into 5 types.
2) \textbf{TCGA-EGFR}: a public dataset with 696 WSIs categorized into 2 types.
3) \textbf{TCGA-LUNG-3K}: a public dataset with 3064 WSIs categorized into 3 types.
Please refer to the supplemental material for detailed information about the datasets.

The WSIs were segmented into non-overlapping patches in size of 224 × 224 under 20× lenses for representation learning and extraction. 
Each dataset was split into training, validation, and testing subsets with the ratio of 6:1:3 at the patient-level.

We followed the implementation in DINO ~\cite{caron2021emerging} to use ViT-S/16 ~\cite{dosovitskiy2020image} as the backbone and the class token of ViT as the final feature embedding.
Our evaluation metrics include average accuracy, micro-average area under the curve (AUC), and macro-average F1 score.
All the methods were implemented in Python 3.8 with torch 1.8.1 and run on a computer cluster with 10 Xeon 2.66GHz CPUs and 10 GPUs of Nvidia Geforce 2080Ti.

\subsection{Comparison with SOTA Methods}
We compared our method with 5 different WSI representing strategies, including ImageNet ~\cite{russakovsky2015imagenet}, SimCLR~\cite{chen2020simple}, BYOL ~\cite{grill2020bootstrap}, MoCov3 ~\cite{Chen_2021_ICCV} and DINO ~\cite{caron2021emerging}.
The first three take ResNet50 ~\cite{he2016deep} as the backbone, while the others utilize ViT-S~\cite{dosovitskiy2020image}.
Moreover, we compared our method with 4 WSI data augmentation frameworks, including DAGAN~\cite{zaffar2022embedding}, ReMix~\cite{yang2022remix}, RankMix~\cite{chen2023rankmix}, Intra-Mixup~\cite{gadermayr2023mixup}.
Given DINO is the basis of our representation learning framework, the comparative experiments with WSI augmentations are performed on the representations extracted by DINO.

As depicted in Figure \ref{fig:chart}, our method achieves the best performance on the USTC-EGFR dataset, TCGA-EGFR dataset and TCGA-LUNG-3K dataset under CLAM~\cite{lu2021data}, TransMIL~\cite{shao2021transmil} and DTFD-MIL~\cite{zhang2022dtfd} benchmarks.

\subsubsection{Comparison with Representation Learning Methods}
SimCLR~\cite{chen2020simple} is the self-supervised representation learning framework widely applied in feature extraction.
The results on the TCGA-EGFR dataset show that it achieves 12.5\%, 2.3\%, 1.9\% AUC better than ImageNet, which showcases the importance of addressing the semantic gap between natural and pathological images.
MoCov3~\cite{Chen_2021_ICCV} and DINO~\cite{caron2021emerging} utilize the ViT architecture as the backbone, demonstrating superior performance on the USTC-EGFR and TCGA-LUNG-3K datasets compared to SimCLR ~\cite{chen2020simple} and BYOL ~\cite{grill2020bootstrap}.
However, the observed performance gaps of MoCov3 and DINO on the TCGA-EGFR dataset compared to top-tier results indicate that current image augmentation strategies may not be sufficiently robust for generating discriminative features needed for effective WSI classification.

Our proposed PRDL additionally introduces representation augmentation into the process of self-supervised learning and employs augmentation prompts to control the representation augmentations, which improves the performance by effectively increasing the diversity of representations.
Compared with our baseline model DINO, PRDL achieves increase in AUC of 7.0\%, 18.4\%, 1.3\% under the CLAM~\cite{lu2021data} benchmark, 6.8\%, 19.2\%, 2.7\% under the TransMIL~\cite{shao2021transmil} benchmark and 5.2\%, 24.4\%, 1.7\% under the DTFD-MIL~\cite{zhang2022dtfd} benchmark on the three datasets,  respectively.
Moreover, the promptable distribution estimator and the augmentation masks trained in representation learning can be utilized in the downstream task.

\subsubsection{Comparison with WSI Augmentation Methods}
Mixup-based methods generate varied representations by mixing features at different levels.
ReMix~\cite{yang2022remix} proposes to mix instance prototypes formed by clustering.
The performance of ReMix in AUC on the TCGA-EGFR dataset has improved by 8.8\% under the CLAM benchmark when compared to DINO.
Nevertheless, the performance degradation observed across all Mixup-based methods on the TCGA-LUNG dataset under the CLAM benchmark suggests that Mixup-based strategies may compromise semantic integrity, leading to noisy training samples and a subsequent decline in model performance.
DAGAN ~\cite{zaffar2022embedding} incorporates a Generative Adversarial Network (GAN) ~\cite{goodfellow2020generative} model to create new representations, adding another training process.
It improves performance in AUC by 0.6\%, 3.8\%, 0.3\% on the USTC-EGFR dataset.
However, the generative model is trained after representation learning so that the performance is limited by the quality of the original representations.
Moreover, DAGAN introduces extra computation costs in both training and inference phases.

Our WSI augmentation strategy PRS, which samples from prompted distributions, offers flexible augmentation, enhancing WSI classification without additional parameters.
This can also be viewed as a reasonable perturbation to representations, which breaks the invariance of representations during training.
With the guide of the augmentation prompts, the noise generated during sampling is much less than Mixup.
Moreover, the distribution estimator is concurrently trained with representation learning, avoiding additional training overhead and continuously adapting to the representation change over the training procedure.
With PRS, there are observed improvements in performance across three datasets, with increases in AUC of 9.4\%, 15.2\%, 2.1\% under the CLAM~\cite{lu2021data} benchmark, 4.8\%, 9.8\%, 1.3\% under the TransMIL~\cite{shao2021transmil} benchmark, 6.6\%, 18.4\%, 1.2\% under the DTFD-MIL~\cite{zhang2022dtfd} benchmark compared to the second-best data augmentation methods.

\begin{table}[t]
    \scriptsize
    \centering
    \setlength{\tabcolsep}{10pt}
    \begin{tabular}{l|cccc}
        \toprule
        Methods & AUC & F1-Score & ACC \\
        \midrule
        DINO & 
        79.4 & 49.8 & 50.2 \\
        PRDL &
        86.4 & 52.5 & 57.9 \\
        PRDL+RS (w/o $\textbf{M}$) &
        86.0 & 59.6 & 59.6 \\
        PRDL+$\textbf{m}_{ResizedCrop}$ &
        89.3 & 61.2 & 61.9 \\
        PRDL+$\textbf{m}_{HorizontalFlip}$ &
        88.8 & 61.0 & 61.0 \\
        PRDL+$\textbf{m}_{ColorJitter}$ &
        88.9 & 61.9 & 62.3 \\
        PRDL+$\textbf{m}_{Grayscale}$ &
        89.3 & 62.1 & 63.2 \\
        PRDL+$\textbf{m}_{GaussianBlur}$ &
        89.0 & 62.7 & 62.8 \\
        PRDL+$\textbf{m}_{Solarization}$ &
        89.1 & 63.0 & 63.2 \\
        \rowcolor[gray]{.8}
        PRDL+PRS &
        \textbf{89.4} & \textbf{65.1} & \textbf{65.9} \\
        \bottomrule
    \end{tabular}
    \caption{Impact of the different augmentation masks on the USTC-EGFR Dataset under the CLAM benchmark.}
    \label{table:ablation2}
\end{table}

\begin{figure}[tb]
  \centering
  \begin{subfigure}{0.8\linewidth}
    \includegraphics[width=\textwidth]{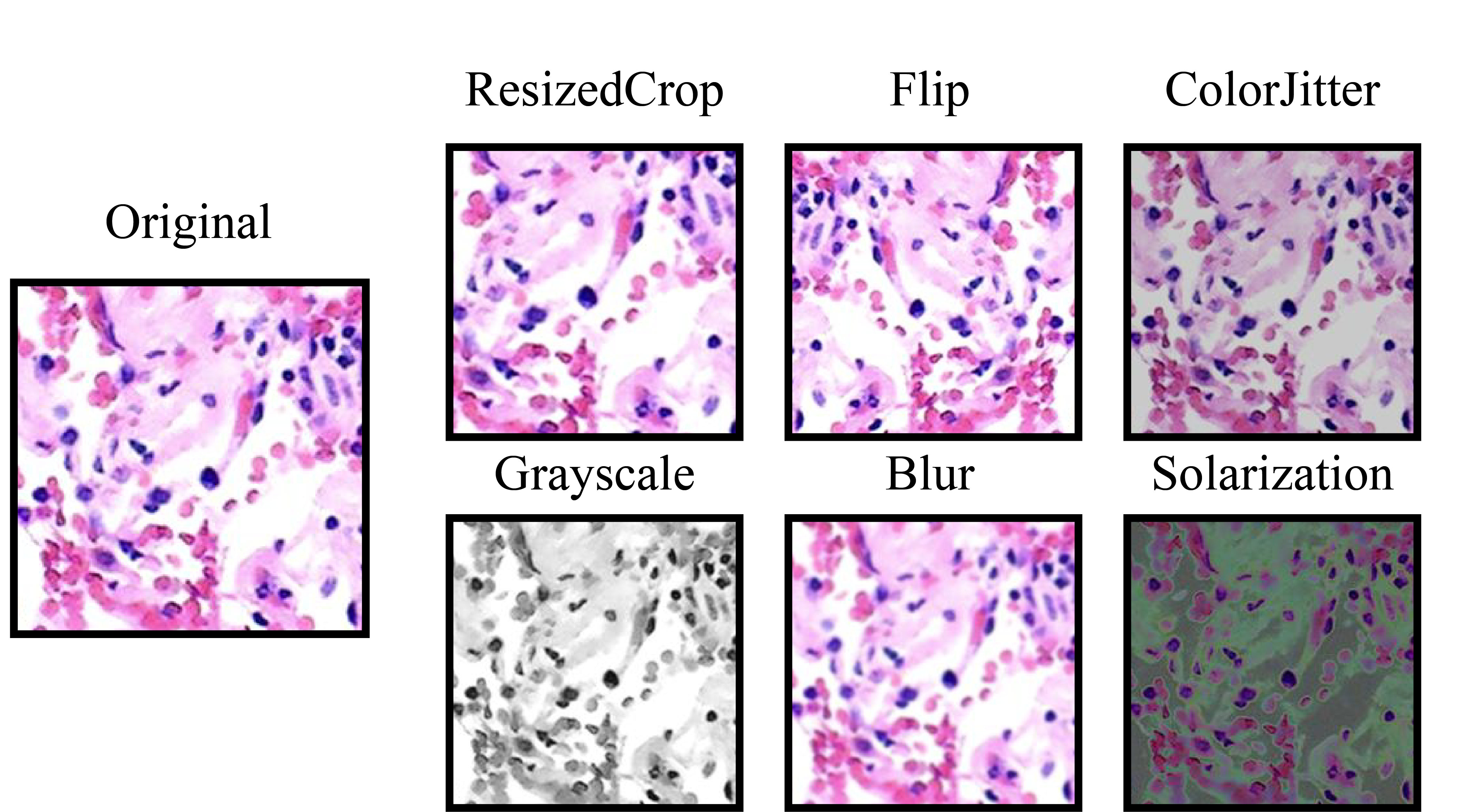}
    \caption{Image Augmentations}
    \label{fig:image_aug}
  \end{subfigure}%
  \hfill 
  \begin{subfigure}{0.85\linewidth}
    \includegraphics[width=\textwidth]{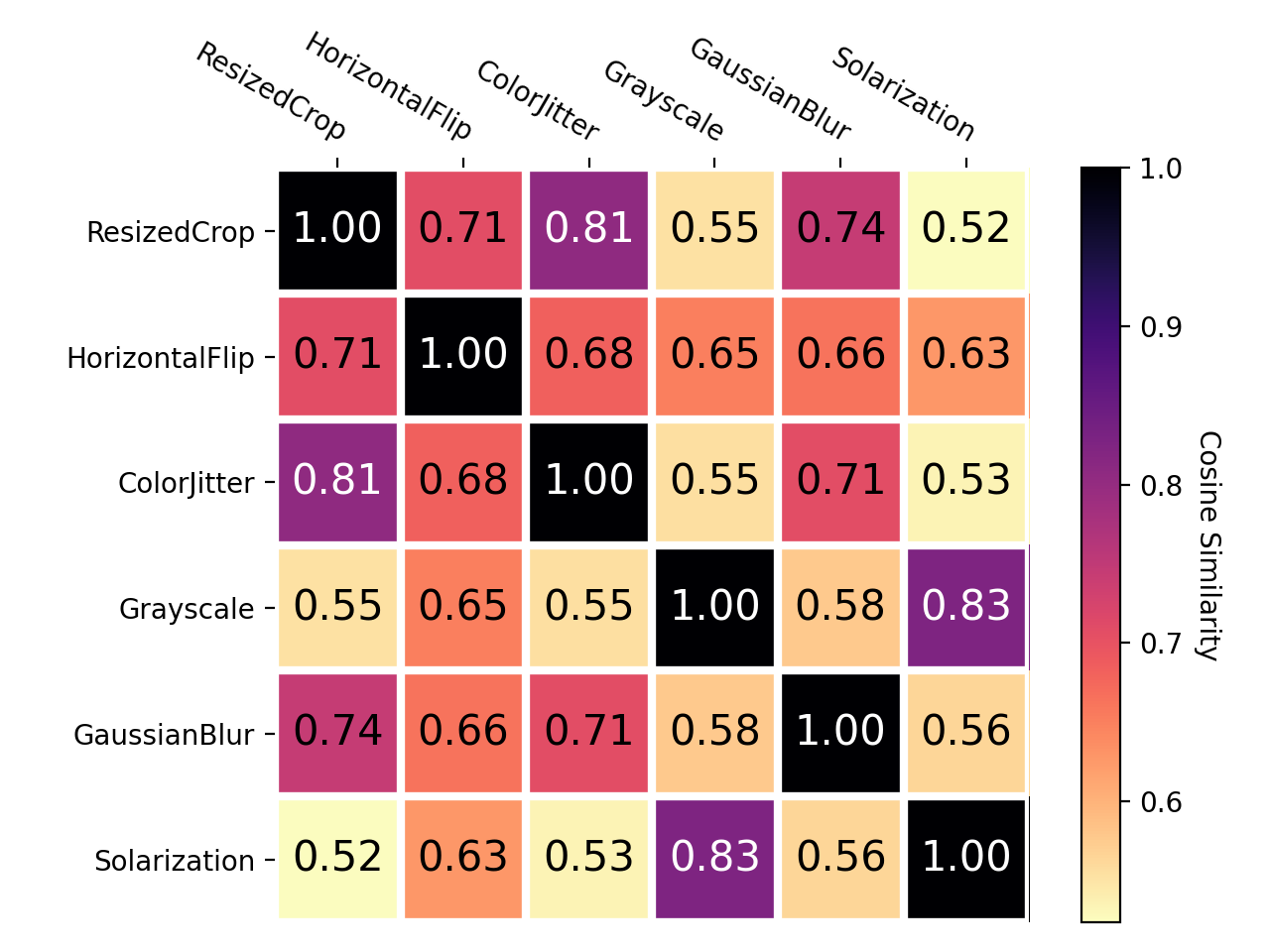}
    \caption{Similarity Matrix}
    \label{fig:sim-egfr}
  \end{subfigure}
  \caption{Dimensional impact between different augmentation prompts on the USTC-EGFR dataset, where (a) is the image augmentations corresponding to the prompts. (b) is the cosine similarities of the augmentation masks \textbf{M}.}
  \label{fig:sim}
\end{figure}

\subsection{Ablation Study}
We first conduct ablation study to verify the necessity of model components. The results are detailed in Table \ref{table:ablation}.
Here, we first investigate WSI augmentation within the feature space through random perturbation using a standard Gaussian distribution.
It shows that the performance on the USTC-EGFR dataset is decreased in F1-Score by 14.8\% compared to DINO under the CLAM benchmark, while it is increased by 1.3\% under the DTFD-MIL benchmark.
This suggests that while perturbations affect the model's performance, they do not completely ruin the representations. 
Meanwhile, we utilize Monte Carlo Sampling ~\cite{zheng2023kernel} where patch representations are randomly discarded during the training of the WSI models. 
As a result, the performance under the DTFD-MIL benchmark is increased in AUC by 5.2\% compared to DINO, which indicates that it works in some cases. 
However, this strategy has a high risk of losing important information for classification, thus leading to significant degradation of performance under the CLAM benchmark.
The results of these two strategies show that the impact of purposeless perturbation on the patch representations is unstable, but opens the possibility of guiding perturbations to enhance the performance of WSI classification.

PRDL w/o PRA denotes PRDL without promptable representation augmentation related operations, where data augmentations are merely carried out in image space.
The decrease in the evaluation metrics demonstrates that the PRA module plays a crucial role in representation learning.
Considering that the representation distribution is estimated based on a Gaussian prior, we constrain it with KL divergence.
The results of PRDL w/o $\mathcal{L}_{KL}$ show that KL divergence is essential for the training of the distribution estimator.
Additionally, components like $\mathcal{L}_{sp}$ and $\mathcal{L}_{var}$ within the PRDL framework are crucial for effective learning of augmentation masks. 
Finally, we verify the impact of the PRS strategy on WSI classification.
We observe that PRS can improve the performance of the WSI classifier even further building on PRDL, where the Micro-AUC under the CLAM benchmark increases from 0.864 to 0.894.
This suggests that the representations generated by PRS are effective to the model and contain useful semantic information. 

\subsection{The impact of augmentation prompts}
As shown in Table \ref{table:ablation2}, we study the impact of different augmentation prompts.
The results of PRDL+RS(w/o $\textbf{M}$) indicate that an overly broad range of representation augmentations can result in limited improvement.
Individual augmentation prompts produce better but varied results, yet they still fall short compared to PRDL+PRS that employs a random combination of the augmentation prompts.
This demonstrates the importance of a guided and variable representation augmentation for improving model performance.
Figure \ref{fig:sim} shows that different augmentation prompts have different effects on the representation dimensions, and some augmentations cause similar impacts in feature space.
\emph{ReisizedCrop} and \emph{HorizontalFlip} are both operated in spatial space and thus their corresponding prompts are similar.
However, \emph{HorizontalFlip} has less improvement in performance compared with other prompts, which indicates that the prompt obtained from this simple image transformation also has relatively limited effects on the augmented representations.
\emph{ColorJitter} and \emph{GaussianBlur} have minimal effects on structural attributes of images, leading to similar impacts in feature space.
Conversely, \emph{Grayscale} and \emph{Solarization} significantly alter the image, which directs focus towards key dimensions in representation, hence their prompts perform better.

\section{Conclusion}
We proposed a novel promptable representation distribution learning (PRDL) framework with a promptable representation sampling (PRS) strategy for promptable and efficient data augmentation in feature space for histopathology WSI classification.
The approach involves a promptable distribution estimator and augmentation prompts for generating diverse representation augmentations, thereby improving the representation quality.
This is complemented by a PRS strategy, tailored to leverage the trained estimator and prompts for effective WSI augmentation.
The prompted representation augmentations significantly enhance image representations and preserve control, thus avoiding the loss of important semantic information, all of which are beneficial for WSI analysis.

\section{Acknowledgments}
This work was partly supported by Beijing Natural Science Foundation (Grant No. 7242270), partly supported by the National Natural Science Foundation of China (Grant No. 62171007, 61901018, and 61906058), partly supported by the Fundamental Research Fund for the Central Universities of China (grant No. YWF-23-Q-1075), partly supported by the Anhui Provincial Natural Science Foundation (Grant
No. 2408085MF162), partly supported by Emergency Key Program of Guangzhou Laboratory (Grant No. EKPG21-32), partly supported by Joint Fund for Medical Artificial Intelligence (Grant No. MAI2023C014), and partly supported by National Key Research and Development Program of China (Grant No. 2021YFF1201004).

\bibliography{aaai25}

\clearpage

\begin{center}
        \centering
        \huge \bfseries Supplementary Material
\end{center}

\section{Dataset Details}
The evaluation of the proposed method involved a private dataset and two public datasets, detailed as below:

\textbf{USTC-EGFR} is a private lung adenocarcinoma dataset that contains 754 WSIs for epidermal growth factor receptor (EGFR) gene mutation identification. 
These WSIs are categorized into 5 classes, including EGFR 19del mutation (19del), EGFR L858R mutation (L858R), none common driver mutations (Wild), other driver gene mutation (Other), and cancer-free tissue (Normal).

\textbf{TCGA-EGFR} is a public lung adenocarcinoma dataset for EGFR mutation classification that contains 696 WSIs collected from the cancer genome atlas (TCGA) program of NCI. 
These WSIs are categorized into 2 classes, including wild type and mutant type.

\textbf{TCGA-LUNG-3K} is a public lung dataset that contains 3064 WSIs collected from the cancer genome atlas (TCGA) program of NCI. 
These WSIs are categorized into 3 classes, including lung adenocarcinoma (LUAD), lung squamous cell carcinoma (LUSC), and cancer-free tissue (Normal).

The detailed distributions of these datasets are shown in Table \ref{table:dataset}.

\section{Training Details}
For the representation learning phase, 1000 patches are sampled from each WSI in the training subset to ensure a comprehensive and diverse set of samples for model learning.
This strategy generates 457,000, 422,000, and 1,839,000 patches, respectively for USTC-EGFR, TCGA-EGFR, and TCGA-LUNG-3K dataset, for the training of the self-supervised models.
Before the training of the WSI classifier begins, each patch is encoded as a Gaussian distribution characterized by mean ($\mathbf{\mu}$) and standard deviation ($\mathbf{\sigma}$).
This process prepares the patches for subsequent classification tasks.

We follow the implementation details of DINO~\cite{caron2021emerging}, to train a backbone of ViT-S/16.
The learning rate is linearly ramped up during warmup with the following linear scaling rule $l_r=0.0005\times N_b/256$ with $N_b$ denoting the batch size.
After warmup, the learning rate and the weight decay follow a cosine schedule.
The temperature $\tau_s$ is set to 0.1 while $\tau_t$ is linearly ramped up from 0.04 to 0.07.
We train the SSL model by 20 epochs on the training set with the batch size as $N_b=32$.
Our data augmentation strategy, detailed in Table \ref{table:aug}, includes three distinct phases. 
Initially, pre-augmentation involves basic spatial and color modifications. 
The subsequent image augmentations add more complex transformations like cropping. 
The student branch mirrors the teacher's operations but incorporates 8 local views.

\section{Hyper-parameter Verification}
\begin{figure*}[tb]
  \centering
  \includegraphics[width=0.8\textwidth]{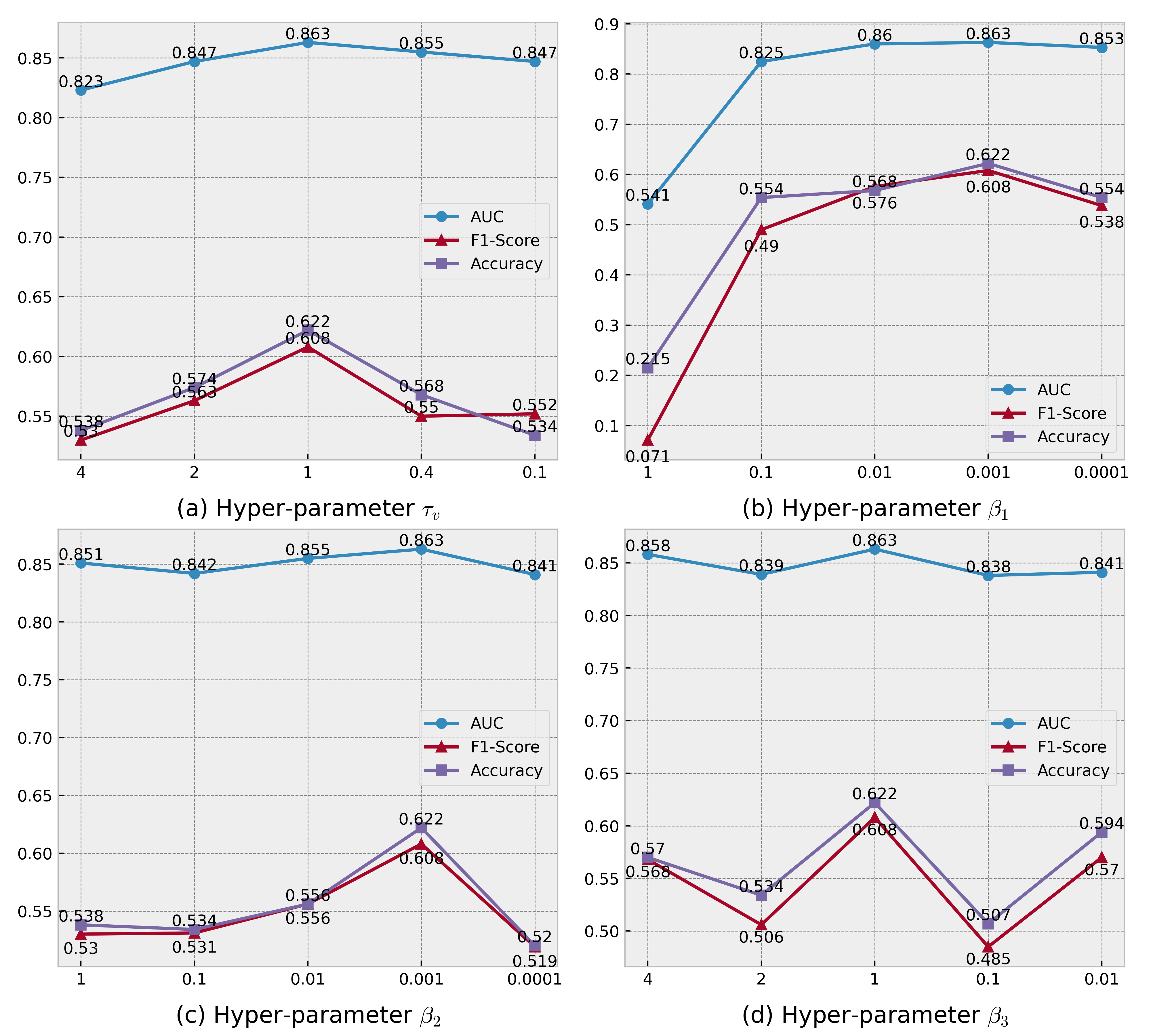}
  \caption{Effects of the hyper-parameters on the USTC-EGFR validation subset under the CLAM benchmark.
  }
  \label{fig:hyper}
\end{figure*}

\begin{table}[tb]
    \scriptsize
    \centering
    \begin{tabular}{l|ccccc}
        \toprule
        \textbf{USTC-EGFR}& Normal& 19del& L858R& Wild& Other \\
        Train& 101& 69& 118& 85& 84 \\
        Val& 16& 11& 19& 14& 14 \\
        Test& 48& 38& 47& 47& 43 \\
        \midrule
        \textbf{TCGA-EGFR}& Wild& Mutant \\
        Train& 357& 65 \\
        Val& 58& 9  \\
        Test& 175& 32 \\
        \midrule
        \textbf{TCGA-LUNG-3K}& Normal& LUAD& LUSC \\
        Train& 339& 740& 760 \\
        Val& 56& 123& 126  \\
        Test& 158& 394& 368 \\
        \bottomrule
    \end{tabular}
    \caption{the WSI Distribution of the Experimental Datasets.}
    \label{table:dataset}
\end{table}

\begin{table}[tb]
    \scriptsize
    \centering
    \begin{tabular}{l|p{4cm}}
        \toprule
        \multirow{3}{*}{Pre-Augmentation} &
        RandomHorizontalFlip \\ & ColorJitter (0.4, 0.4, 0.2, 0.1) \\ & RandomGrayscale \\
        \midrule
        \multirow{5}{*}{Image Augmentation Teacher} &
        RandomResizedCrop (224) \\ & RandomHorizontalFlip \\ & ColorJitter (0.4, 0.4, 0.2, 0.1) \\ &
        GaussianBlur \\ & Solarization \\
        \midrule
        \multirow{6}{*}{Image Augmentation Student} &
        RandomResizedCrop (224) \\ & RandomResizedCrop (96) \\ & RandomHorizontalFlip \\ & 
        ColorJitter (0.4, 0.4, 0.2, 0.1) \\ & GaussianBlur \\ & Solarization \\
        \bottomrule
    \end{tabular}
    \caption{Composition of Different Augmentations.}
    \label{table:aug}
\end{table}

The main factors that decide the capacity of the PRDL include 4 hyper-parameters $(\tau_v, \beta_1, \beta_2, \beta_3)$. 
We recorded Micro-AUC, F1 score, and accuracy metrics for the validation subset to assess the impact of each hyper-parameter while keeping other hyper-parameters constant.
The detailed outcomes of this tuning process are illustrated in Figure \ref{fig:hyper}.
\subsubsection{The temperature of representation augmentation}
The sharpness of representations from the representation augmentation student branch is regulated by adjusting the softmax temperature parameter $\tau_v$.
Optimal performance is observed with lower temperatures. 
As the temperature increases, the targets become smoother and the evaluation metrics decrease.
Consequently, we have set $\tau_v=1$ for subsequent experiments.
\subsubsection{The weight of KL divergence}
The hyper-parameter $\beta_1$ controls the coefficient of $\mathcal{L}_{KL}$, which constrains the distribution estimator in the PRDL framework.
We observe that from the performance on the validation subset is improved with a decrease in $\beta_1$.
Therefore, we have set $\beta_1=0.001$ for PRDL training to optimize performance.
\subsubsection{The weight of L1 normalization}
The hyper-parameter $\beta_2$ sets the weight of $\mathcal{L}_{sp}$, impacting the sparsity of augmentation masks in PRDL training.
The optimal performance on the validation subset is achieved at $\beta_2=0.001$, hence this value has been chosen for training the PRDL to enhance performance.
\subsubsection{The weight of variance regularization}
The hyper-parameter $\beta_3$ determines the weight of $\mathcal{L}_{var}$ that affects the variance of augmentation masks across the feature dimension during PRDL training.
$\beta_3=1$ is found to achieve the best performance on the validation subset, thus we have select this value in PRDL training.

\section{Complete Numerical Results}
The complete numerical results for Fig.3 in the body of the paper are summarized in Table \ref{table:egfr}, \ref{table:tcga-egfr}, \ref{table:lung}.

\begin{table}[t]
    \scriptsize
    \centering
    \setlength{\tabcolsep}{5pt}
    \begin{tabular}{l|cccc}
        \toprule
        Methods & USTC-EGFR & TCGA-EGFR & TCGA-LUNG-3K \\
        \midrule
        CONCH~\cite{lu2024avisionlanguage} &
        85.0 & 65.4 & 96.0 \\
        \rowcolor[gray]{.8}
        PRDL &
        86.4 & 81.4 & \textbf{97.9} \\
        \rowcolor[gray]{.8}
        PRDL+PRS &
        \textbf{89.4} & \textbf{87.0} & 97.5 \\
        \bottomrule
    \end{tabular}
    \caption{AUC comparison with the foundation model under the CLAM benchmark.}
    \label{table:foundation}
\end{table}

\begin{table}[t]
    \scriptsize
    \centering
    \setlength{\tabcolsep}{3pt}
    \begin{tabular}{l|cccc}
        \toprule
        Methods & USTC-EGFR & TCGA-EGFR & TCGA-LUNG-3K \\
        \midrule
        DINO+PAMA~\cite{wu2024pan} &
        79.8 & 70.7 & 96.6\\
        \rowcolor[gray]{.8}
        PRDL+PRS+PAMA &
        \textbf{85.9} & \textbf{85.8} & \textbf{98.0} \\
        \bottomrule
    \end{tabular}
    \caption{AUC comparison under the PAMA framework.}
    \label{table:pama}
\end{table}

\section{Comparison with the Foundation Model}
 Given the large foundation models in histopathology~\cite{lu2024avisionlanguage} have demonstrated robust representational capabilities, we evaluated the performance of the foundation models across the datasets under the CLAM benchmark. The AUC results are detailed in Table \ref{table:foundation}. It shows that CONCH~\cite{lu2024avisionlanguage} performs well in the TCGA-LUNG-3K dataset, which demonstrates its good generalization in common cancer subtyping task. However, the AUC of CONCH is 21.6\% lower than our method in the TCGA-EGFR dataset. It indicates that well-designed representation pre-training and augmentation methodologies are still necessary for challenging prediction tasks. Moreover, our method is not specifically designed for certain cancers and is general for histopathology patch representation learning, which makes it adequate to build histopathology foundation models.

\section{Integration with WSI-level Pretraining}
WSI-level pretraining is one of the downstream tasks that benefits from data augmentation techniques~\cite{wu2024pan}.
Our method can enhance the variability of the data augmentation strategies used in WSI-level pertaining, thereby improving the generalization of the model.
We experiment with a fine-tuning training strategy that involves training all network parameters after pretraining.
The results in Table \ref{table:pama} show that our method achieves increase in AUC of 6.1\%, 15.1\% and 1.4\% on the three datasets.
This demonstrates that our method can work effectively in conjunction with the state-of-the-art WSI-level pretraining methods.

\begin{table*}[!p]
    \footnotesize
    \centering
    \begin{tabular}{l|ccc|ccc|ccc}
        \toprule
        \multirow{2}{*}{\textbf{Methods}} & \multicolumn{3}{c|}{\textbf{CLAM}~\cite{lu2021data}} & \multicolumn{3}{c|}{\textbf{TransMIL}~\cite{shao2021transmil}}& \multicolumn{3}{c}{\textbf{DTFD-MIL}~\cite{zhang2022dtfd}}\\ 
        & AUC & F1-Score & ACC &
        AUC & F1-Score & ACC &
        AUC & F1-Score & ACC \\
        \midrule
        ImageNet~\cite{russakovsky2015imagenet} & 
        77.9 & 38.4 & 43.1 &
        73.4 & 30.4 & 34.5 &
        73.1 & 38.3 & 39.0 \\
        SimCLR~\cite{chen2020simple} &
        69.4 & 38.1 & 39.5 &
        62.5 & 22.8 & 30.9 &
        70.3 & 37.8 & 40.4 \\
        BYOL~\cite{grill2020bootstrap} &
        61.5 & 25.6 & 31.8 &
        54.8 & 26.0 & 20.7 &
        61.7 & 28.7 & 34.1 \\
        MoCov3~\cite{Chen_2021_ICCV} &
        77.6 & 40.7 & 43.5 &
        63.9 & 23.1 & 29.6 &
        74.7 & 31.9 & 38.1 \\
        DINO~\cite{caron2021emerging} &
        79.4 & 49.8 & 50.2 &
        75.2 & 41.3 & 44.0 &
        76.1 & 41.5 & 43.1 \\
        \midrule
        DINO+Random Perturbation &
        74.9 & 35.0 & 40.8 &
        74.1 & 39.2 & 39.9 &
        75.6 & 42.8 & 43.9 \\
        DINO+MC Sampling~\cite{zheng2023kernel} &
        75.9 & 37.2 & 39.5 &
        75.2 & 36.7 & 38.6 &
        81.3 & 45.0 & 47.1 \\
        DINO+ReMix~\cite{yang2022remix} &
        78.2 & 37.1 & 43.5 &
        75.6 & 36.3 & 39.9 &
        73.7 & 36.7 & 41.7 \\
        DINO+RankMix~\cite{chen2023rankmix} &
        73.3 & 29.7 & 32.7 &
        75.5 & 36.7 & 43.1 &
        73.9 & 30.4 & 35.9 \\
        DINO+Intra-Mixup~\cite{gadermayr2023mixup} &
        79.0 & 38.7 & 40.8 &
        79.7 & 42.1 & 43.9 &
        76.8 & 41.8 & 42.6 \\
        DINO+DAGAN~\cite{zaffar2022embedding} &
        80.0 & 49.0 & 51.1 &
        79.0 & 42.6 & 44.4 &
        76.4 & 44.9 & 46.6 \\
        DINO+AugDiff~\cite{dai2024augdiff} &
        81.7 & 48.8 & 52.3 &
        77.1 & 35.2 & 39.8 &
        75.5 & 42.5 & 43.8 \\
        \midrule
        \rowcolor[gray]{.8}
        PRDL &
        86.4 & 52.5 & 57.9 &
        82.0 & 47.1 & 48.4 &
        81.3 & 48.7 & 49.8 \\
        \rowcolor[gray]{.8}
        PRDL+PRS &
        \textbf{89.4} & \textbf{65.1} & \textbf{65.9} &
        \textbf{84.5} & \textbf{55.1} & \textbf{56.5} &
        \textbf{83.4} & \textbf{51.3} & \textbf{52.9} \\
        \bottomrule
    \end{tabular}
    \caption{Comparisons with SOTA Methods on the USTC-EGFR Dataset.}
    \label{table:egfr}
    \begin{tabular}{l|ccc|ccc|ccc}
        \toprule
        \multirow{2}{*}{\textbf{Methods}} & \multicolumn{3}{c|}{\textbf{CLAM}~\cite{lu2021data}} & \multicolumn{3}{c|}{\textbf{TransMIL}~\cite{shao2021transmil}}& \multicolumn{3}{c}{\textbf{DTFD-MIL}~\cite{zhang2022dtfd}}\\
        & AUC & F1-Score & ACC &
        AUC & F1-Score & ACC &
        AUC & F1-Score & ACC \\
        \midrule
        ImageNet~\cite{russakovsky2015imagenet} & 
        63.8 & 61.3 & 76.8 &
        65.5 & 54.6 & 75.4 & 
        73.4 & 62.4 & 76.3 \\
        SimCLR~\cite{chen2020simple} &
        76.3 & 61.7 & 73.9 &
        67.8 & 60.1 & 70.5 &
        75.3 & 60.1 & 79.6 \\
        BYOL~\cite{grill2020bootstrap} &
        60.9 & 51.3 & 78.3 &
        64.0 & 56.0 & 70.1 &
        58.9 & 55.6 & 76.8 \\
        MoCov3~\cite{Chen_2021_ICCV} &
        48.1 & 45.7 & 68.1 &
        58.2 & 49.0 & 70.1 &
        63.6 & 55.9 & 78.7 \\
        DINO~\cite{caron2021emerging} &
        63.0 & 54.6 & 75.4 &
        59.1 & 55.1 & 82.2 &
        59.4 & 53.1 & 74.9 \\
        \midrule
        DINO+Random Perturbation &
        61.5 & 53.4 & 64.7 &
        63.4 & 57.4 & 71.0 &
        67.6 & 52.4 & 59.9 \\
        DINO+MC Sampling~\cite{zheng2023kernel} &
        70.7 & 60.5 & 77.8 &
        66.9 & 60.5 & 76.8 &
        69.0 & 54.2 & 80.2 \\
        DINO+ReMix~\cite{yang2022remix} &
        71.8 & 47.4 & 81.6 &
        61.2 & 45.8 & 84.5 &
        58.7 & 50.7 & 74.9 \\
        DINO+RankMix~\cite{chen2023rankmix} &
        61.8 & 36.5 & 57.4 &
        62.9 & 36.5 & 57.4 &
        63.9 & 45.8 & 84.5 \\
        DINO+Intra-Mixup~\cite{gadermayr2023mixup} &
        67.9 & 63.1 & 64.7 &
        73.6 & 68.5 & 69.1 &
        69.5 & 59.3 & 74.3 \\
        DINO+DAGAN~\cite{zaffar2022embedding} &
        71.0 & 39.6 & 52.9 &
        71.7 & 64.4 & 64.7 &
        61.6 & 57.8 & 76.8 \\
        DINO+AugDiff~\cite{dai2024augdiff} &
        67.6 & 60.6 & 77.3 &
        67.8 & 54.5 & 78.8 &
        73.4 & 61.8 & 80.8 \\
        \midrule
        \rowcolor[gray]{.8}
        PRDL &
        81.4 & 70.6 & \textbf{84.1} &
        78.3 & 67.9 & 80.2 &
        83.8 & 71.6 & 85.5 \\
        \rowcolor[gray]{.8}
        PRDL+PRS &
        \textbf{87.0} & \textbf{75.1} & 83.6 &
        \textbf{83.4} & \textbf{77.0} & \textbf{86.5} &
        \textbf{87.9} & \textbf{77.2} & \textbf{85.5} \\
        \bottomrule
    \end{tabular}
    \caption{Comparisons with SOTA Methods on the TCGA-EGFR Dataset.}
    \label{table:tcga-egfr}
    \begin{tabular}{l|ccc|ccc|ccc}
        \toprule
        \multirow{2}{*}{\textbf{Methods}} & \multicolumn{3}{c|}{\textbf{CLAM}~\cite{lu2021data}} & \multicolumn{3}{c|}{\textbf{TransMIL}~\cite{shao2021transmil}}& \multicolumn{3}{c}{\textbf{DTFD-MIL}~\cite{zhang2022dtfd}}\\
        & AUC & F1-Score & ACC &
        AUC & F1-Score & ACC &
        AUC & F1-Score & ACC \\
        \midrule
        ImageNet~\cite{russakovsky2015imagenet} & 
        94.6 & 84.1 & 82.6 &
        91.9 & 80.2 & 78.2 &
        93.7 & 82.8 & 80.9 \\
        SimCLR~\cite{chen2020simple} &
        93.5 & 80.3 & 78.8 &
        91.9 & 81.2 & 79.0 &
        93.0 & 82.5 & 81.0 \\
        BYOL~\cite{grill2020bootstrap} &
        92.6 & 78.5 & 77.1 &
        89.6 & 75.9 & 74.6 &
        90.9 & 78.3 & 76.9 \\
        MoCov3~\cite{Chen_2021_ICCV} &
        94.1 & 82.6 & 81.4 &
        93.1 & 79.7 & 77.7 &
        93.9 & 82.3 & 80.4 \\
        DINO~\cite{caron2021emerging} &
        96.6 & 87.2 & 85.8 &
        94.7 & 85.5 & 83.6 &
        95.7 & 87.4 & 85.5 \\
        \midrule
        DINO+Random Perturbation &
        96.4 & 86.7 & 85.2 &
        95.0 & 84.9 & 83.3 &
        96.1 & 87.2 & 85.4 \\
        DINO+MC Sampling~\cite{zheng2023kernel} &
        96.1 & 87.2 & 85.7 &
        95.2 & 84.1 & 83.0 &
        96.3 & 87.9 & 86.6 \\
        DINO+ReMix~\cite{yang2022remix} &
        95.4 & 83.8 & 82.8 &
        93.7 & 83.0 & 81.3 &
        96.1 & 84.4 & 81.8 \\
        DINO+RankMix~\cite{chen2023rankmix} &
        94.1 & 81.6 & 79.5 &
        93.4 & 75.3 & 73.2 &
        94.2 & 75.7 & 73.8 \\
        DINO+Intra-Mixup~\cite{gadermayr2023mixup} &
        94.6 & 80.9 & 80.6 &
        96.0 & 87.7 & 86.1 &
        96.6 & 89.4 & 88.0 \\
        DINO+DAGAN~\cite{zaffar2022embedding} &
        94.1 & 81.6 & 79.5 &
        94.7 & 85.4 & 83.6 &
        95.5 & 86.0 & 84.2 \\
        DINO+AugDiff~\cite{dai2024augdiff} &
        95.9 & 86.4 & 85.1 &
        94.5 & 84.2 & 82.7 &
        95.8 & 87.4 & 85.8 \\
        \midrule
        \rowcolor[gray]{.8}
        PRDL &
        \textbf{97.9}& 90.5& 89.1 &
        \textbf{97.4}& 89.8& 88.5 &
        97.4& 90.3& 89.0 \\
        \rowcolor[gray]{.8}
        PRDL+PRS &
        97.5& \textbf{90.6}& \textbf{89.2}& 
        97.3& \textbf{90.5}& \textbf{89.4}&
        \textbf{97.8}& \textbf{91.1}& \textbf{89.9} \\
        \bottomrule
    \end{tabular}
    \caption{Comparisons with SOTA Methods on the TCGA-LUNG-3K Dataset.}
    \label{table:lung}
\end{table*}

\end{document}